\documentclass{article}

\usepackage[final,nonatbib]{neurips_2023}

\usepackage{color}
\usepackage{amsmath}
\usepackage{amssymb}
\usepackage{amsbsy}
\usepackage{adjustbox}
\usepackage{subfigure}
\usepackage{array}
\usepackage{colortbl}
\usepackage{arydshln}
\usepackage{graphicx}
\usepackage{dashrule}
\usepackage{fontawesome}
\usepackage{multirow}
\usepackage{pifont}
\usepackage{tikz}
\usetikzlibrary{arrows.meta}

\usepackage[utf8]{inputenc} % allow utf-8 input
\usepackage[T1]{fontenc}    % use 8-bit T1 fonts
\usepackage{hyperref}       % hyperlinks
\usepackage{url}            % simple URL typesetting
\usepackage{booktabs}       % professional-quality tables
\usepackage{amsfonts}       % blackboard math symbols
\usepackage{nicefrac}       % compact symbols for 1/2, etc.
\usepackage{microtype}      % microtypography
\usepackage{xcolor}         % colors

\definecolor{skyblue}{RGB}{23,194,217}

\definecolor{darkgreen}{rgb}{0.0, 0.5, 0.0}
\definecolor{lightblue}{RGB}{173,216,230}

\newcommand{\pinkcircleicon}{%
\begin{tikzpicture}[scale=0.5]
    \fill[pink!40] (0,0) circle (0.3);
\end{tikzpicture}%
}
\newcommand{\bluecircleicon}{%
\begin{tikzpicture}[scale=0.5]
    \fill[lightblue!30] (0,0) circle (0.3);
\end{tikzpicture}%
}
\newcommand{\greencircleicon}{%
\begin{tikzpicture}[scale=0.5]
    \fill[green!15] (0,0) circle (0.3);
\end{tikzpicture}%
}
\newcommand{\skybluecircleicon}{%
\begin{tikzpicture}[scale=0.5]
    \fill[skyblue] (0,0) circle (0.3);
\end{tikzpicture}%
}

\newcommand{\greycircleicon}{
\begin{tikzpicture}[scale=0.5]
    \fill[gray!30, draw=black] (0,0) circle (0.3);
\end{tikzpicture}
}

\newcommand{\myreddashedarrow}{
\begin{tikzpicture}
\draw [red, dashed, -{Latex}] (0,0) -- (1,0);
\end{tikzpicture}
}
\title{Towards Zero Memory Footprint Spiking Neural Network Training}

\author{%
Bin Lei$^{1}$ \quad Sheng Lin \quad \textbf{Pei-Hung Lin}$^2$ \quad \textbf{Chunhua Liao}$^2$ \quad \textbf{Caiwen Ding}$^1$\\
$^1$University of Connecticut \quad  $^2$Lawrence Livermore National Laboratory \quad \\
\texttt{\{bin.lei,caiwen.ding\}@uconn.edu} \texttt{\{shenglin4227\}@gmail.com}\\
\texttt{\{lin32,liao6\}@llnl.gov}
}

\hypersetup{
    colorlinks=true,
    linkcolor=red,
    citecolor=blue,
    urlcolor=blue
}

\begin{document}

\maketitle

\begin{abstract}
Biologically-inspired Spiking Neural Networks (SNNs),  processing information using discrete-time events known as spikes rather than continuous values, have garnered significant attention due to their hardware-friendly and energy-efficient characteristics.  
However, the training of SNNs necessitates a considerably large memory footprint, given the additional storage requirements for spikes or events, leading to a complex structure and dynamic setup.
In this paper, to address memory constraint in SNN training, we introduce an innovative framework, characterized by a remarkably low memory footprint. We \textbf{(i)} design a reversible SNN node that retains a high level of accuracy. Our design is able to achieve a $\mathbf{58.65\times}$ reduction in memory usage compared to the current SNN node. We \textbf{(ii)} propose a unique algorithm to streamline the backpropagation process of our reversible SNN node. This significantly trims the backward Floating Point Operations Per Second (FLOPs), thereby accelerating the training process in comparison to current reversible layer backpropagation method. By using our algorithm, the training time is able to be curtailed by $\mathbf{23.8\%}$ relative to existing reversible layer architectures.
\end{abstract}

\section{Introduction}

% \begin{figure*}[!h]
% %\vspace{-.1in}
% % \begin{center}
% \hspace*{0.5cm}
% \includegraphics[width=1\linewidth]{figures/shpere.pdf}
% \caption{Reversible Spiking Neural Networks are more memory-efficient. The size of the data points are proportional to the number of paramaters.}
% %  \vspace{-3mm}
% \label{fig:intro}
% % \end{center}
% \end{figure*}
% The computational complexity and energy consumption of deep learning models present challenges~\cite{kugele2020efficient,shrestha2022survey}. 
% This has led to a new computing paradigm: energy-efficient, bio-inspired neuromorphic computing.
As 
% an  energy-efficient,
a bio-inspired neuromorphic computing representative, Spiking Neural Network (SNN) has attracted considerable attention, in contrast to the high computational complexity and energy consumption of traditional Deep Neural Networks (DNNs)~\cite{tang2019bridging,davies2021advancing,shrestha2022survey,han2020deep}. 
% In recent years, Spiking Neural Networks (SNNs) have gained significant attention as a powerful computational paradigm that closely resembles the functionality of biological neurons.
SNN
%  discrete spike events
% rather than continuous values, which helps reduce hardware design complexity
% are a type of artificial neural network that 
processes information using discrete-time events known as spikes rather than continuous values, offering extremely hardware-friendly and energy-efficient characteristics. For instance, in a robot navigation task using Intel's Loihi~\cite{davies2018loihi}, SNN could achieve a 276$\times$ reduction in energy compared to a conventional DNN approach. Work ~\cite{rueckauer2022nxtf} shows that DNN consumes 111mJ and 1035mJ per sample on MNIST and CIFAR-10, respectively, while SNN consumes only 0.66mJ and 102mJ, i.e., 168$\times$  and 10$\times$  energy reduction. 

% SNN consumes only 0.66mJ and 102mJ per sample on MNIST and CIFAR-10, respectively, while DNN consumes 111mJ and 1035mJ, i.e., 168$\times$  and 10$\times$  energy reduction. 

% The applications of SNNs are broad and varied, spanning across domains such as Robotics and Control Systems~\cite{hulea2020bioinspired}, Auditory Processing~\cite{bos2023sub}, and Computer Vision~\cite{datta2023sensor}. SNNs outshine traditional Convolutional Neural Networks (CNNs) in several ways: they excel at processing temporal data~\cite{mohamed2019review}, closely mimic biological neuron behavior~\cite{tang2019bridging}, and potentially offer superior power efficiency~\cite{han2020deep}. 
% Firstly, SNNs better mimic biological neural systems as they use spikes to convey information, which is more akin to the behavior of real neurons~\cite{tang2019bridging}. Secondly, due to their asynchronous and event-driven nature, they excel in handling time-series data and quickly responding to events~\cite{mohamed2019review}. In addition, SNNs have high energy efficiency, as they consume energy only when the neurons fire. This makes SNNs tremendously promising in terms of energy efficiency, especially for scenarios with limited power resources such as edge computing and mobile devices~\cite{han2020deep}.
% \textcolor{red}{(not exciting. Why SNN is better than CNN? in terms of what? )}
% \todo{Leo: What successful use cases have been reported for SNNs?}

Despite their numerous advantages, one major bottleneck in the deployment of SNNs has been memory consumption.
% , resulting in a large memory footprint. 
The memory complexity of a traditional DNN with a depth of $L$ is $\mathcal{O}(L)$. But for SNN of the same depth $L$, there are several timesteps $T$ involved in the computation. Consequently, the memory complexity of SNNs escalates to $\mathcal{O}(L*T)$. 
% This memory demands of SNNs will increase due to the incorporation of temporal dynamics. 
For instance, the memory requirement during the DNN training process of ResNet19 is 0.6 GB, but for the SNN with the same architecture could reach about 12.34 GB (\textasciitilde20 $\times$) when time-step equals 10.
% \todo{Leo: how large? Are the space complexity notation like O(??) for SNNs compared to other types of Neurual networks?} 
This presents a significant challenge for their applicability to resource-constrained systems, such as IoT-Edge devices~\cite{putra2021q}.

To address the issue of high memory usage of SNN, researchers have proposed several methods, including Quantization~\cite{putra2021q}, Weight Sparsification~\cite{rathi2018stdp,huang2023neurogenesis}, the Lottery Ticket Hypothesis~\cite{kim2022lottery}, Knowledge Distillation~\cite{guo2023joint}, Efficient Checkpointing~\cite{singh2022skipper}, and so on.
In this paper, we introduce a novel reversible SNN node that drastically compresses the memory footprint of the SNN node inside the entire network. Our method achieves state-of-the-art results in terms of SNN memory savings. It achieves this by recalculating all the intermediate states on-the-fly, rather than storing them during the backward propagation process. To further enhance the efficiency of our approach, we also present a new algorithm for the backpropagation process of our reversible SNN node, which significantly reduces the training time compared with the original reversible layer backpropagation method. Remarkably, our method maintains the same level of accuracy throughout the process.

% Contribution
% \\ \todo{exetramly low memory footprint and activation value free checkpointing mechnizem.}\\
% \\ \todo{computation efficient backprogration}
% \\ \todo{lossless}
% \todo{Leo: what is the new time complexity notation compared to traditional ones?} 
% \todo{How many-fold} \todo{Leo: what is the new space complexity notation O(??)?} 
% \todo{average accuracy loss in how many tasks}
% As a result of our innovations, We reduce the memory complexity of the SNN node from $\mathcal{O}(n^2)$ to $\mathcal{O}(1)$. 
% , while maintaining comparable accuracy to traditional SNNs. Moreover, our method reduces the FLOPs needed for the backpropagation by a factor of 23\% compared to existing reversible architecture, thus accelerate the training process.
% Collectively, these advances pave the way for more efficient and scalable SNN implementations, enabling the deployment of these biologically inspired networks across a wider range of applications and hardware platforms.
As a result of our innovations, we reduce the memory complexity of the SNN node from $\mathcal{O}(n^2)$ to $\mathcal{O}(1)$, while maintaining comparable accuracy to traditional SNN node. Moreover, our method reduces the FLOPs needed for the backpropagation by a factor of 23\% compared to existing reversible layer backpropagation method, thus accelerating the training process. Collectively, these advances pave the way for more efficient and scalable SNN implementations, enabling the deployment of these biologically inspired networks across a wider range of applications and hardware platforms.

\section{Background And Related Works} 

\subsection{Spiking Neural Network}
% \todo{Ask Shaoyi to change 2.1}
% \todo{Add loss function. Show the normal backpropagation}
% \newpage
% As one of the most promising strategy for implementing low-power neuromorphic hardware, 
Spiking neural network (SNN) uses sparse binary spikes over multiple time steps to deal with visual input in an event-driven manner ~\cite{davies2018loihi, diehl2015unsupervised}. We use SNN with the popular Leaky Integrate and Fire (LIF) spiking neuron. The forward pass is formulated as follows.

% \vspace{-0.5cm}
% \begin{subequations}
% \label{eq:neuron_model}
% \begin{gather}
%         v[t] = \alpha v[t{-}1] + \sum_i w_i s_i[t]  - \vartheta o[t-1] \label{eq:voltage} \\
%         o[t]  = u(v[t] - \vartheta)  \label{eq:threshold} \\
%         h(x) = 0, x < 0 \text{ otherwise 1} \label{eq:heaviside}
% \end{gather}
% \end{subequations}
% \vspace{-0.2cm}
\vspace{-0.5cm}
\begin{subequations}
\label{eq:neuron_model}
\begin{gather}
        v[t] = \alpha v[t{-}1] + \sum_i w_i s_i[t]  - \vartheta o[t-1] \label{eq:voltage} \\
        o[t]  = h(v[t] - \vartheta)  \label{eq:threshold} \\
        h(x) = \begin{cases} 
           0, & \text{if } x < 0 \\
           1, & \text{otherwise}
           \end{cases}
        \label{eq:heaviside}
\end{gather}
\end{subequations}
\vspace{-0.2cm}

where $t$ denotes time step. In Eq.~\eqref{eq:voltage}, $v[t]$ is the dynamics of the neuron's membrane potential after the trigger of a spike at time step t. 
% $\alpha \in (0,1]$ controls the $v[t]$ decay speed. 
The sequence $s_i[t] \in \{0,1\}$ represents the $i$-th input spike train, consisting solely of 0s and 1s, while $w_i$ is the corresponding weight. In Eq.~\eqref{eq:threshold}, $o[t] \in \{0,1\}$ is the neuron's output spike train. In Eq.~\eqref{eq:heaviside}, $h(x)$ is the Heaviside step function to generate the outputs.

In the backward pass, we adopt Backpropagation Through Time (BPTT) to train SNNs. The BPTT for SNNs using a surrogate gradient~\cite{neftci2019surrogate}, which is formulated as follows. 
% \todo{Leo: surrogate is used because the unit step function is not differentiable?}

\begin{subequations}
\label{eq:neuron_model}
\begin{gather}
    \delta_l[t] = \epsilon_{l+1}[t] w_{l+1} \label{eq:error}\\
    \epsilon_{l}[t] = \delta_{l}[t] \phi_{l}[t] + \alpha \epsilon_{l}[t] 
    \label{eq:epsilon}\\
    \frac{\partial L}{\partial w_l} = \sum_{t=0}^{T-1} \epsilon_l[t] \cdot [s_l[t]]^\intercal \label{eq:grad}
    \end{gather}
\end{subequations}

Denote $L$ as the loss, in Eq.~\eqref{eq:error}, $\delta_l[t] {=} \frac{\partial L}{\partial o_l[t]}$ is the error signal at layer $l$ time step $t$ which is propagated using above formulations. In Eq.~\eqref{eq:epsilon}, $\epsilon_l[t] {=} \frac{\partial L}{\partial v_l[t]}$, $\phi_l[t] {=} \frac{\partial o_l[t]}{\partial v_l[t]} {=} \frac{\partial u(v_l[t] - \vartheta)}{\partial v_l[t]}$, where we follow the gradient surrogate function in~\cite{fang2021deep} to approximate the derivative of $u(x)$, such that $\frac{\partial u(x)}{\partial x} \approx \frac{1}{1 + \pi^2 x^2}$.
Eq.~\eqref{eq:grad} calculates the gradient of $l^{th}$ layer weight $w_l$.

\subsection{Reversible Layer}

A reversible layer refers to a family of neural network architectures that are based on the Non-linear Independent Components Estimation~\cite{dinh2014nice,dinh2016density}. 
% During the backpropagation process, gradients need to be computed through every layer of the entire network. 
In traditional training methods, the activation values of each layer are stored in memory, which leads to a dramatic increase in memory requirements as the network depth increases. However, reversible transformation technology allows us to store only the final output of the network and to recompute the discarded ones when needed. This approach significantly reduces memory requirements, making it possible to train deeper neural networks and more complex models under limited memory conditions, thus potentially unlocking new insights and improving performance across a wide range of tasks. The reversible transformation has been used in different kinds of neural networks, such as CNN~\cite{gomez2017reversible}, Graph neural networks (GNN)~\cite{li2021training}, Recurrent Neural Networks (RNN)~\cite{mackay2018reversible} and some transformers~\cite{mangalam2022reversible}. Furthermore, studies have demonstrated the effectiveness of reversible layers in different types of tasks, including image segmentation~\cite{pendse2021memory}, natural language processing~\cite{kitaev2020reformer}, compression~\cite{liu2021semantics}, and denoising~\cite{huang2022winnet}. 
% Its widespread adoption demonstrates the significant role it plays in enabling the training of deeper and more complex models under memory constraints, ultimately pushing the boundaries of what is achievable in deep learning. 
% \todo{Leo: if space permits, a diagram showing the revsersible layer may be helpful.}
\section{Reversible SNN Algorithm}
% \newcommand{\blackcircle}[1]{%
%     \tikz{%
%         \node[draw, circle, fill=black, text=white, inner sep=1pt, minimum size=1.2em] (num) {#1};
%     }%
% }
% Define a new command for the grey circle icon
% \newcommand{\greycircleicon}{%
% \begin{tikzpicture}[scale=0.5]
%     \fill[gray!30, draw=black] (0,0) circle (0.3);
% \end{tikzpicture}%
% }

% \newcommand{\reddashedarrow}{
% \begin{tikzpicture}
% \draw [red, dashed, -latex] (0,0) -- (1,0);
% \end{tikzpicture}
% }

%%%%%%%%%%%%%%%%%%%%%%%%%%%%%%%%%%%%%%%reversible SNN function%%%%%%%%%%%%%%%%%%%%%%%%%
\subsection{Reversible SNN Memory Analysis} 
\label{memory analysis}
\begin{figure}[htbp]
    \centering
    % \hspace*{-2cm}
    \includegraphics[width=1\textwidth]{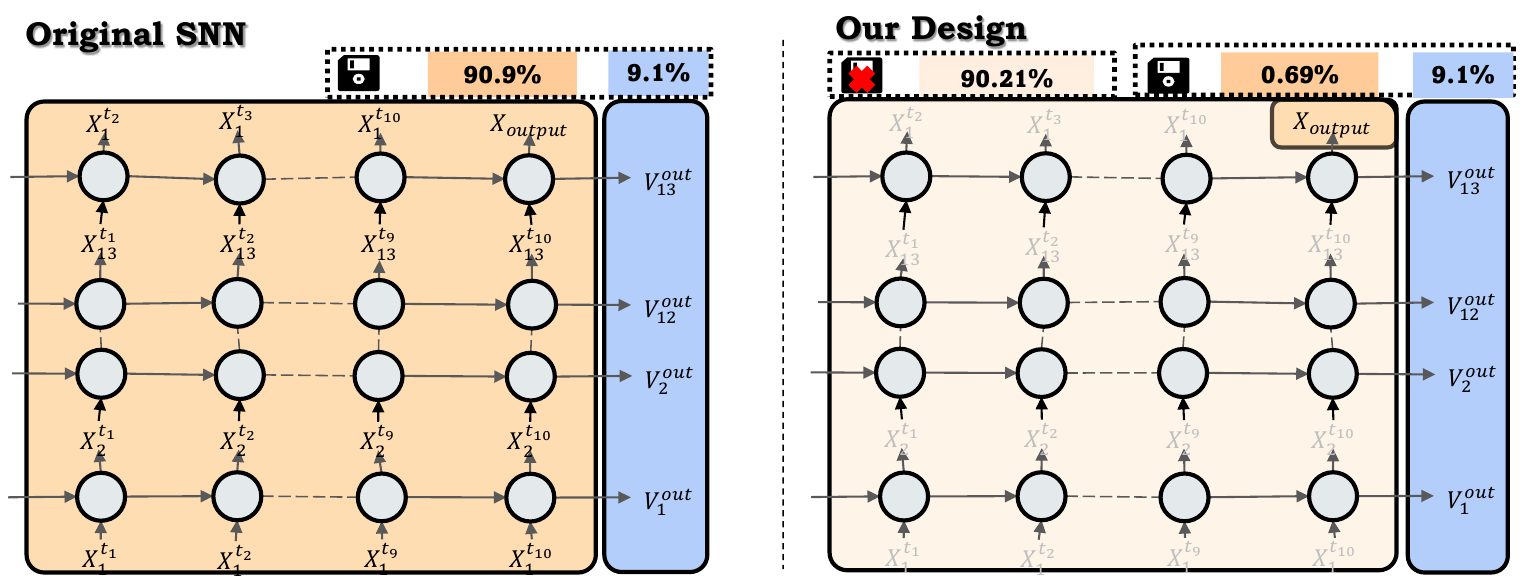} 
    \caption{Activation value Memory Comparison between the original SNN network and our reversible SNN network, using VGG13 with a timestep of ten as an example.~\greycircleicon: SNN node,~\faSave: Save to memory,~\faSave~with~{\color{red}\ding{55}}: NOT save to memory.}
    \label{fig:SNN memory}
\end{figure}

During the training process of SNN networks, the activation values occupy the main memory storage space. The SNN activation value memory analysis schematic diagram is shown in Fig.\ref{fig:SNN memory}. 
%%%%%%%%%%%%%%%%%%%%%%%%%%%%%%%%%%%%%%%%Memory_saving_figure%%%%%%%%%%%%%%%%%%%%%%%%
% \todo{Leo: how to read the diagram? What do different icons on the top mean? It may be better to show absolute memory footprints for each, then show the saving percentage.}
In this figure, we use the VGG-13 architecture~\cite{simonyan2014very} with ten timesteps as an example. The percentage values represent the memory footprint ratio of each part in the entire network. The left diagram is the original SNN where the activation values of $X$ account for 90.9\% of the memory usage, and the output potentials of each neuron occupy 9.1\% of the memory. The right diagram is our designed reversible SNN, which only requires saving the final $X_{output}$ values and the output potentials of each neuron, without storing all intermediate values, thus significantly saving memory. The intermediate activation values will be regained during the backpropagation process through our inverse calculation equation. In this example, our method is able to save 90.21\% of the memory used for activation values. The exact amount of memory saved by our method will be shown in Experiment part. 
%%%%%%%%%%%%%%%%%%%%%%%%%%%%%%%%%%%%%%%Forward subsection%%%%%%%%%%%%%%%%%%%%%%%%%
\subsection{Reversible SNN Forward Calculation} 
\label{sec:Reversible SNN forward calculation}
\begin{figure}[htbp]
    \centering
    \includegraphics[width=1\textwidth]{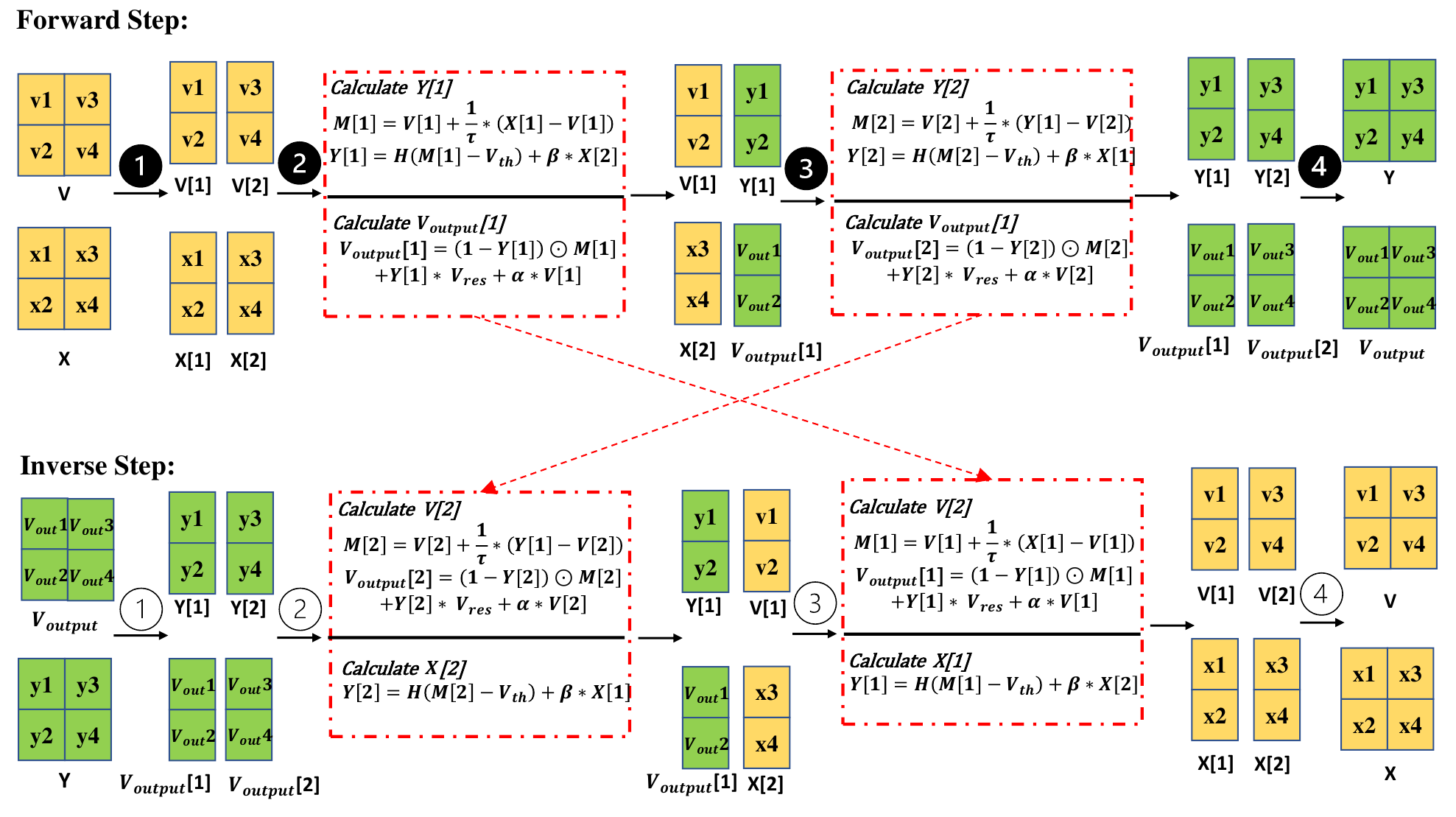} 
    \caption{This reversibility demo use $2 \times 2$ toy Input as an example and shows our forward and inverse calculations.~\protect\myreddashedarrow: 
 The origin of the equations in the inverse process.}
    \label{fig:reversible rnn}
\end{figure}

Our forward algorithm is in the upper section of Fig.\ref{fig:reversible rnn}. 
%%%%%%%%%%%%%%%%%%%%%%%%%%%%%%%%%%%%%%%%Toy_example_Figure%%%%%%%%%%%%%%%%%%%%%%%%
% \todo{Leo: the crossed red arrows are a bit counter-intuitive. What do they mean? }
\ding{182}: The various input states $S = (X, V)$ of each neuron are evenly divided into two groups along the last dimension. Namely: $S = [S_1, S_2]$. 
% Consider $S$ with shape $(m, n, 2p)$, $\forall~0 \leq i < m$, $0 \leq j < n$, $0 \leq k < p$, we have $S_1(i, j, k) = S(i, j, k)$, $S_2(i, j, k) = S(i, j, p + k)$. Hence, $S = [S_1, S_2]$.

\ding{183}: Calculate the first part of output $Y_1$:

% \noindent
% \begin{minipage}{0.45\linewidth}
% \begin{adjustbox}{valign=c, margin = 3em 0em 0em 0em}
% \begin{equation}
% \label{eq:forward M0}
% M_1^t = V_1^{t-1} + \frac{1}{\tau} \cdot\left(X_1^t - V_1^{t-1}\right) \hspace{2em}
% \end{equation}
% \end{adjustbox}
% \end{minipage}
% \hfill
% \begin{minipage}{0.45\linewidth}
% \begin{adjustbox}{valign=c}
% \begin{equation}
% Y_1^{t} = H\left(M_1^{t} - V_{th}\right) + \beta \cdot X_2^t \hspace{2em}
% \label{eq:forward Y0}
% \end{equation}
% \end{adjustbox}
% \end{minipage}

\begin{minipage}[c]{0.45\linewidth}
    \begin{equation}
        M_1^t = V_1^{t-1} + \frac{1}{\tau} \cdot\left(X_1^t - V_1^{t-1}\right)
        \label{eq:forward M0}
    \end{equation}
\end{minipage}
\hfill
\begin{minipage}[c]{0.45\linewidth}
    \begin{equation}
        Y_1^{t} = H\left(M_1^{t} - V_{th}\right) + \beta \cdot X_2^t
        \label{eq:forward Y0}
    \end{equation}
\end{minipage}

$M_1^t$ is the membrane potential of the first half neuron at time $t$.
$V_2^{t-1}$ is the input potential of the second half neuron at time $t-1$.
$\tau$ is the time constant.
$X_2^t$ is the input to the second half neuron at time $t$.
$V_{th}$ is the threshold voltage of the neurons.
$H()$ is the Heaviside step function.
$\beta$ is a scaling factor for the input. $\beta \cdot X_2^t$ will help $Y_1^{t}$ to collect information about the second half of the input in the next step. Then calculate first part of output voltage $V_1^t$: 
\setlength{\abovedisplayskip}{8pt}
\setlength{\belowdisplayskip}{8pt}
\begin{equation}
\label{eq:forward V0}
V_1^t=\left(1-Y_1^{t}\right) \odot M_1^t+Y_1^{t} \cdot V_{res}+\alpha \cdot V_1^{t-1}
\end{equation}
\setlength{\abovedisplayskip}{8pt}
\setlength{\belowdisplayskip}{8pt}
$V_1^t$ is the output potential of the first half neuron at time $t$.
$V_{res}$ is the reset voltage of the neurons.
$\alpha$ is a scaling factor for the membrane potential.

\ding{184}: Use the first part of output $Y_1$ to calculate the second part $Y_2$:

% \noindent
% \begin{minipage}{0.45\linewidth}
% \begin{adjustbox}{valign=c, margin = 3em 0em 0em 0em}
% \begin{equation}
% \label{eq:forward M1}
%  M_2^t=V_2^{t-1}+\frac{1}{\tau}\left(Y_1^{t}-V_2^{t-1}\right) \hspace{2em}
% \end{equation}
% \end{adjustbox}
% \end{minipage}
% \hfill
% \begin{minipage}{0.45\linewidth}
% \begin{adjustbox}{valign=c}
% \begin{equation}
% \label{eq:forward Y1}
% Y_2^{t}=H\left(M_2^t - V_{th}\right)+\beta \cdot X_1^t \hspace{2em}
% \end{equation}
% \end{adjustbox}
% \end{minipage}

\noindent
\begin{minipage}[c]{0.45\linewidth}
    \begin{equation}
        M_2^t = V_2^{t-1} + \frac{1}{\tau}\left(Y_1^{t} - V_2^{t-1}\right)
        \label{eq:forward M1}
    \end{equation}
\end{minipage}
\hfill
\begin{minipage}[c]{0.45\linewidth}
    \begin{equation}
        Y_2^{t} = H\left(M_2^t - V_{th}\right) + \beta \cdot X_1^t
        \label{eq:forward Y1}
    \end{equation}
\end{minipage}

$M_2^t$ is the membrane potential of the second half neuron at time $t$.
$Y_2^t$ is the output of the second half neuron at time $t$.
calculate the second part of output voltage $V_2^t$:
\setlength{\abovedisplayskip}{8pt}
\setlength{\belowdisplayskip}{8pt}
\begin{equation}
\label{eq:forward V1}
V_2^t=\left(1-Y_2^{t}\right) \odot M_2^t+Y_2^{t} \cdot V_{res}+\alpha \cdot V_2^{t-1}
\end{equation}
\setlength{\abovedisplayskip}{8pt}
\setlength{\belowdisplayskip}{8pt}
$V_2^t$ is the output potential of the second half neuron at time $t$.

\ding{185}: For all the output states $S_{output}=([Y_1,Y_2],[V_1^t,V_2^t])$, combine them by the last dimension.
%%%%%%%%%%%%%%%%%%%%%%%%%%%%%%%%%%%%%%%%inverse subsection%%%%%%%%%%%%%%%%%%%%%%%%%%%%%%%%%%%%%%%
\subsection{Reversible SNN Inverse Calculation}
The purpose of the inverse calculation is to use the output results to obtain the unsaved input values. i.e. Use $Y$ and $V_{output}$ to calculate $X$ and $V$. Our inverse algorithm is in the lower section of Fig.\ref{fig:reversible rnn}.

\ding{192}: For all the output states $S_{output}=(Y,V_{output})$, divide them into two groups by the last dimension in the same way as in the first step of forward calculation, namely: $S_{output}=[S_{output}1;S_{output}2]$

\ding{193}: Calculate $V_1^t$ by combine Eq.\ref{eq:forward M1} and Eq.\ref{eq:forward V1}, simplify:
\setlength{\abovedisplayskip}{8pt}
\setlength{\belowdisplayskip}{8pt}
\begin{equation}
\label{eq:inverse V1}
V_2^{t-1} = \frac{V_2^t - (1 - Y_2) \cdot \frac{1}{\tau} \odot Y_1 - Y_2 \cdot V_{reset}}{(1 - Y_2) \cdot (1 - \frac{1}{\tau}) + \alpha}
\end{equation}
\setlength{\abovedisplayskip}{8pt}
\setlength{\belowdisplayskip}{8pt}
Calculate $X_1^t$ by combine Eq.\ref{eq:forward M1} and \ref{eq:forward Y1}, simplify:
\setlength{\abovedisplayskip}{8pt}
\setlength{\belowdisplayskip}{8pt}
\begin{equation}
\label{eq:inverse X0}
X_1^{t}=\left(Y_2^{t} - H\left(M_2^t - V_{th}\right)\right)\div\beta
\end{equation}
\setlength{\abovedisplayskip}{8pt}
\setlength{\belowdisplayskip}{8pt}
\ding{194}: Calculate $V_1^t$ by combine Eq.\ref{eq:forward M0} and Eq.\ref{eq:forward V0}, simplify:
\setlength{\abovedisplayskip}{8pt}
\setlength{\belowdisplayskip}{8pt}
\begin{equation}
\label{eq:inverse V1}
V_1^{t-1} = \frac{V_1^t - (1 - Y_1) \cdot \frac{1}{\tau} \odot X_1^{t} - Y_1 \cdot V_{reset}}{(1 - Y_1) \cdot (1 - \frac{1}{\tau}) + \alpha}
\end{equation}
\setlength{\abovedisplayskip}{8pt}
\setlength{\belowdisplayskip}{8pt}
Calculate $X_1^t$ by combine Eq.\ref{eq:forward M0} and \ref{eq:forward Y0}, simplify:
\setlength{\abovedisplayskip}{8pt}
\setlength{\belowdisplayskip}{8pt}
\begin{equation}
\label{eq:inverse X0}
X_2^{t}=\left(Y_1^{t} - H\left(M_1^t - V_{th}\right)\right)\div\beta
\end{equation}
\setlength{\abovedisplayskip}{8pt}
\setlength{\belowdisplayskip}{8pt}
\ding{195}: For all the input states $S=([X_1,X_2],[V_1^{t-1},V_2^{t-1}])$, combine them by the last dimension.

\section{Inverse Gradient Calculation}
\label{Inverse calculation of gradients}

Although our reversible architecture significantly reduces memory usage, it does extend computation time for two primary reasons: (i) It necessitates the recalculation of the activation values that weren't originally stored. (ii) Many of the previous reversible layer architectures have inherited the backpropagation method from checkpointing~\cite{cogdl2023,fan2020pyslowfast}. This method requires using the recalculated intermediate activation values to rerun the forward equation, thereby constructing a forward computational graph. This graph is then used to derive the corresponding gradients. This step of rerunning the forward equation introduces additional computational overhead, which extends the overall computation time. 

\begin{figure}[htbp]
    % \centering
    % \hspace*{-1.5cm}
    \includegraphics[width=0.99\textwidth]{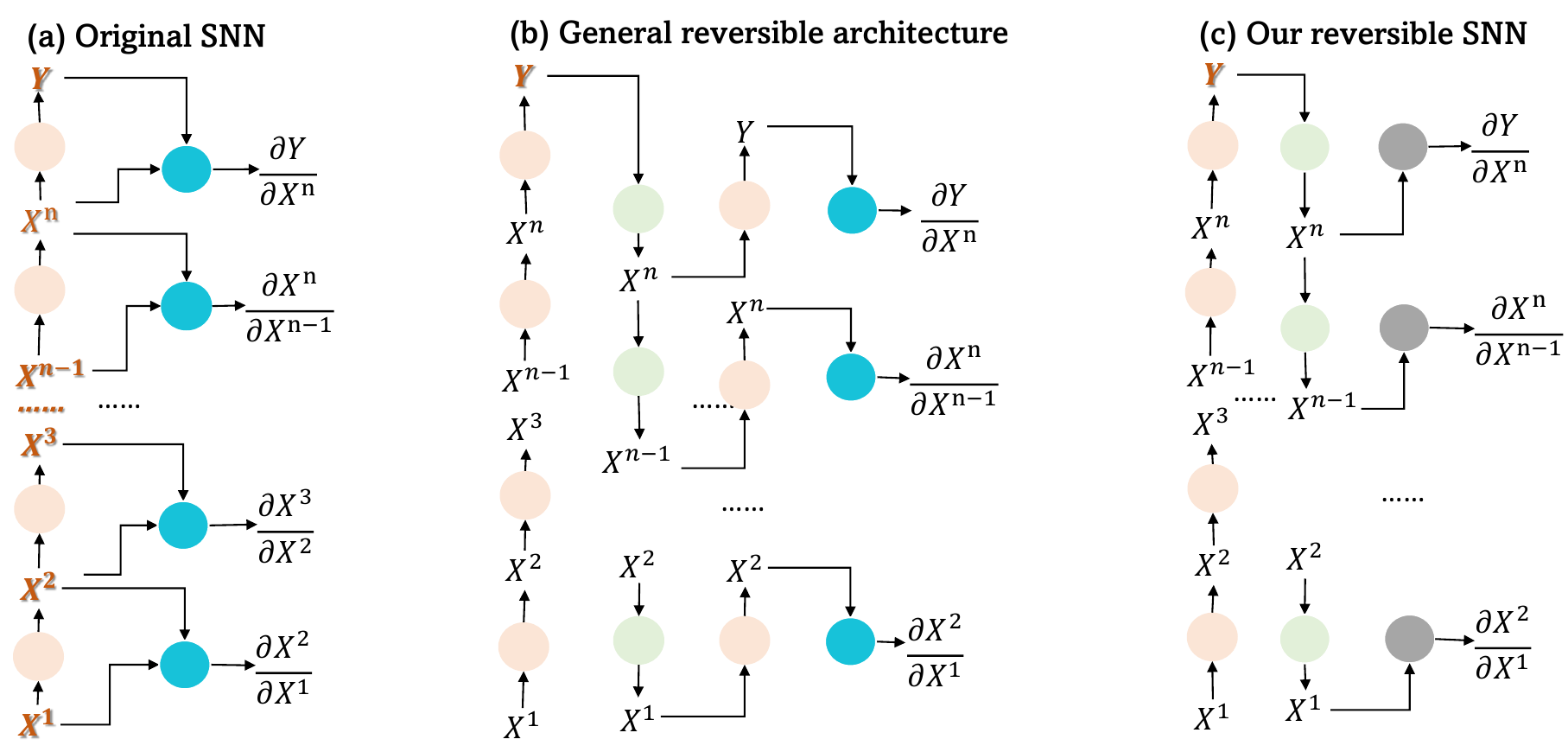} 
    \caption{Three different architectures for comparison.~\protect\pinkcircleicon: Forward function,~\protect\greencircleicon: inverse function,~\protect\skybluecircleicon: $\frac{\partial \mathbf{X^n}}{\partial \mathbf{X^{n-1}}}$ derivative,~\protect\bluecircleicon:\textbf{Part of} $\frac{\partial \mathbf{X^{n-1}}}{\partial \mathbf{X^{n}}}$ derivative,~\textcolor{brown}{\textbf{Brown}} variables: Cached values.}
    \label{fig:inverse gradient}
\end{figure}

This scenario is prevalent across all existing architectures of reversible layers, including Reversible GNN~\cite{li2021training}, Reversible RNN~\cite{mackay2018reversible}, Reversible Transformers~\cite{mangalam2022reversible}, and so on. To reduce the training time, we have designed a new algorithm, called the inverse gradient calculation method, which is able to substantially decrease the number of FLOPs during the backpropagation process compared to the original reversible architecture. Our design is shown in Fig.\ref{fig:inverse gradient}.

The left diagram illustrates the original forward and backward processes. The middle diagram depicts the original calculation process for reversible layers. It contains four steps:
\begin{enumerate}
  \item The input $X$ pass the forward function to compute the output $Y$, without storing the input data to conserve memory.
  \item For each layer $n$: The output $X^n$ of this layer pass the inverse function to compute the input $X^{n-1}$ of this layer. This process starts with the final output $Y$.
  % and updates it through in-place modifications, which do not consume any additional memory.
  \item For each layer $n$: The input $X^{n-1}$ passes through the forward function again to reconstruct the forward computational graph, which facilitates gradient computation.
  \item For each layer $n$: Compute the gradient $\frac{\partial \mathbf{X^n}}{\partial \mathbf{X^{n-1}}}$ based on the forward computational graph.
\end{enumerate}
The right diagram is our design, it contains three steps:
\begin{enumerate}
  \item The input $X$ pass the forward function to compute the output $Y$, without storing the input data to conserve memory.
  \item For each layer $n$: The output $X^n$ of this layer pass the inverse function to compute the input $X^{n-1}$ of this layer and construct an inverse computational graph.
  \item For each layer $n$: Compute the gradient $\frac{\partial \mathbf{X^{n}}}{\partial \mathbf{X^{n-1}}}$ based on the inverse computational graph.
\end{enumerate}

% In this paper, we have streamlined the process by omitting the last two steps and directly constructing a reverse computational graph in the second step. Utilizing this graph, we compute the forward derivative of matrix $\mathbf{Y}$ with respect to matrix $\mathbf{X}$. In other words, we aim to solve the following mathematical problem: Given the partial derivatives of matrix $\mathbf{X}$ with respect to matrix $\mathbf{Y}$, denoted as $\frac{\partial \mathbf{X}}{\partial \mathbf{Y}}$, determine the partial derivatives of matrix $\mathbf{Y}$ with respect to matrix $\mathbf{X}$, denoted as $\frac{\partial \mathbf{Y}}{\partial \mathbf{X}}$.

% Under normal circumstances, reversible layers can be viewed as a trade-off between time and space in computational algorithms. One drawback of this approach is the relatively high computational cost involved. However, by employing the method of computing a reverse matrix graph, the demand for floating-point operations (FLOPs) can be significantly reduced. The following is a detailed estimation of the FLOPs required:

Below is the specific calculation formula of the $\frac{\partial \mathbf{X^{n}}}{\partial \mathbf{X^{n-1}}}$ based on the inverse computation graph, and the derivation process is in the Appendix. 
% \todo{Leo: quite some math formulas are involved. Are there any tools to double check all the math computation steps are correct? There should be some tools to compute derivatives of functions for example.}

% \begin{equation}
% \label{eq:Inverse_dY/dx_1}
% \frac{\partial \mathbf{X^{n}} }{\partial \mathbf{X_1^{n-1}}}=\frac{\theta}{2 + \left(\pi \cdot \theta \cdot \left(M_1^t - V_{th}\right)\right)^2} \cdot \frac{1}{\tau} + \frac{\theta}{2 + \left(\pi \cdot \theta \cdot \left(M_2^t - V_{th}\right)\right)^2} \odot \frac{\theta}{2 + \left(\pi \cdot \theta \cdot \left(M_1^t - V_{th}\right)\right)^2} \cdot \frac{1}{\tau^2} + \beta
% \end{equation}
\begin{equation}
\label{eq:Inverse_dY/dx_1}
\frac{\partial \mathbf{X^{n}} }{\partial \mathbf{X_1^{n-1}}}=\frac{\theta}{2 + \left(\pi \cdot \theta \cdot \left(M_1^t - V_{th}\right)\right)^2} \cdot \frac{1}{\tau} \odot \left(1 + \frac{\theta}{2 + \left(\pi \cdot \theta \cdot \left(M_2^t - V_{th}\right)\right)^2} \cdot \frac{1}{\tau}\right) + \beta
\end{equation}

\begin{equation}
\label{eq:Inverse_dY/dx_2}
\frac{\partial \mathbf{X^{n}} }{\partial \mathbf{X_2^{n-1}}}=\frac{\theta}{2 + \left(\pi \cdot \theta \cdot \left(M_2^t - V_{th}\right)\right)^2} + \beta
\end{equation}

% \begin{equation}
% \label{eq:Inverse_dY/dx_1}
% \frac{\partial \mathbf{Y} }{\partial \mathbf{X_1}}=\beta-\frac{1}{\beta} \cdot \frac{1}{\tau} \cdot \frac{\partial \mathbf{X_2}}{\partial \mathbf{Y_1}} \odot\left(\frac{1}{\tau} \cdot \beta+\frac{\tau-1}{\tau}\left(-\beta \cdot \mathbf{M_1}+\left(1-\mathbf{Y_1}\right)\cdot V_{res}\right)\right)
% \end{equation}
% %8

% \begin{equation}
% \label{eq:Inverse_dY/dx_2}
% \begin{aligned}
% & \frac{\partial \mathbf{Y} }{\partial \mathbf{X_2}}=\frac{\alpha \cdot \left(1-\beta \frac{\partial \mathbf{X_1}}{\partial \mathbf{Y_1}}\right)}{\mathbf{Y_2^t}-1}+\left(\beta-\frac{1}{\beta} \cdot \frac{1}{\tau} \frac{\partial \mathbf{X_2}}{\partial \mathbf{Y_1}} \odot\left(\frac{\alpha(1-\beta\frac{\partial \mathbf{X_1}}{\partial \mathbf{Y_1}})}{\tau(\mathbf{Y_2^t}-1)}\right.\right. \\
% &\left.+\frac{\tau-1}{\tau} \cdot\left(-\frac{\alpha \cdot \left(1-\beta \frac{\partial \mathbf{X_1}}{\partial \mathbf{Y_1}}\right)}{\mathbf{Y_2^t}-1} \odot \mathbf{M_1}+\frac{1-\mathbf{Y_1}}{\tau}\right)\right)
% \end{aligned}
% \end{equation}
% %12

All the variables in Eq.\ref{eq:Inverse_dY/dx_1} and Eq.\ref{eq:Inverse_dY/dx_2} have the same meaning as the variables in Eq.\ref{eq:forward M0}-Eq.\ref{eq:inverse X0} and $\theta$ is an adjustable constant parameter.

The ability to perform computational graph inverse computation in my algorithm is based on that our forward function has symmetry with the inverse computation function.

For the original reversible network:
\setlength{\abovedisplayskip}{8pt}
\setlength{\belowdisplayskip}{8pt}
\begin{equation}
% \label{eq:forward V0}
FLOPS_{backward}^{ori} = FLOPS_{inverse} + FLOPS_{forward} + FLOPS_{\frac{\partial \mathbf{X^n}}{\partial \mathbf{X^{n-1}}}}
\end{equation}
\setlength{\abovedisplayskip}{8pt}
\setlength{\belowdisplayskip}{8pt}
For our reversible network:
\setlength{\abovedisplayskip}{8pt}
\setlength{\belowdisplayskip}{8pt}
\begin{equation}
% \label{eq:forward V0}
FLOPS_{backward}^{our} = FLOPS_{inverse} + FLOPS_{part~of \frac{\partial \mathbf{X^{n-1}}}{\partial \mathbf{X^{n}}}}
\end{equation}
\setlength{\abovedisplayskip}{8pt}
\setlength{\belowdisplayskip}{8pt}

Compared to the standard reversible network, our method reduces FLOPS by 23\%. The FLOPS analysis is shown in Appendix and the detailed time 
% \todo{is this called time mearsurement?} 
mearsurement is shown in the Experiment part.
% By using pytorch's own derivative function, we can easily deduce $\frac{\partial X_1}{\partial Y_1}$ and $\frac{\partial X_2}{\partial Y_1}$ when $\frac{\partial X}{\partial Y}$ is known. After finish calculating the $\frac{\partial \mathbf{Y} }{\partial \mathbf{X_1}}$ and $\frac{\partial Y}{\partial X_2}$, we can get the value of $\frac{\partial Y}{\partial X}$. 

% We obtain the gradient of the forward derivative perfectly by the Inverse calculation of computation gradients, thus eliminating the FLOPs needed for the second forward steps and the normal gradient calculation steps inside the normal reversible layer calculation steps(The short red arrow and short blue arrow in the middle figure of \ref{fig:inverse gradient}), \textcolor{red}{By using our methods, the calculation time saved a lot.} The back propagation process reduces FLOPs by roughly 23\% compared to the normal reversible layer algorithm, see the appendix for the exact estimation process.

% \todo{insert specific FLOPS estimation}

% \todo{Try C++ engine}

% During the forward process, all intermediate values are not saved, and the inverse function is employed to recompute the activation values from the final layer. Subsequently, the activation values are passed through the forward function again to construct the computation graph. Finally, the backward process is used to calculate the gradient.
% \subsection{Memory Savings Analysis}
\section{Experiment}
% We carried out a comprehensive set of experiments on CIFAR-10, \textcolor{red}{CIFAR-100}~\cite{krizhevsky2009learning}, and Tiny-ImageNet~\cite{chrabaszcz2017downsampled} datasets, employing various network architectures such as VGG11, VGG13, VGG16, VGG19, ResNet19, ResNet34, ResNet50~\cite{he2016deep}, and ResNet101 for our experimental setup. 
We first compare our design with the SOTA SNN Memory-efficient methods for SNN training on the several datasets.
Subsequently, we incorporate our reversible SNN node into different architectures across various datasets. 
% \todo{Leo: correctly implementing the math formulas in Python maybe nontrivial and error-prone. How do you ensure your code is correct? Is the final model's accuracy a sufficient proof in the field for correctness?} 
The goal is twofold: Firstly, we wanted to demonstrate that compared to the SNN node currently in use, our reversible version is able to offer substantial memory savings. Secondly, we aim to show that, when compared to the existing reversible layer backpropagation method, our reversible SNN node backpropagation design is able to considerably reduce the time spent in the backpropagation process, thereby accelerating the training phase.
In the final part, we conduct an ablation study to evaluate the influence of various parameters within our equations and the impact of the number of groups into which the input is divided on the performance of our model.

All experiments were conducted on a Quadro RTX6000 GPU equipped with 24GB of memory, using PyTorch 1.13.1 with CUDA 11.4, and an Intel(R) Xeon(R) Gold 6244 CPU running at 3.60GHz. To ensure that the values we obtain through inverse-calculation is the same as the original forward-calculation method, we use torch.allclose(rtol=$1e^{-06}$, atol=$1e^{-10}$) to compare all the inverse calculated values with the origianl forward calculated values. All the results return true as they should be. Detailed hyperparameter settings for each experiment are provided in the Appendix.
\subsection{Comparison with the SOTA Methods} 
\label{Comparison with the SOTA Methods}
We conducted a comparison of our approach with the current SOTA methods in memory efficiency during the SNN training process on the CIFAR10 and CIFAR100 datasets. To verify the universality of our work, we apply our designed reversible SNN node to the current SOTA sparse training work for SNNs. A comparison was then made between these two methods on the Tiny-ImageNet dataset, the results are shown in Table~\ref{tab:ComparisonOfSNNMethods}.

\begin{table}[htbp]
  \centering
  \caption{Comparison of our work with the SOTA methods in memory efficiency during the SNN training process. For all the works: Batch size = 128. \small{{*}: They did not provide the memory data directly for training CIFAR100, we estimate it based on their memory usage for training CIFAR10 and their parameter data.}}
  \scalebox{0.8}{
    \begin{tabular}{cccccc}
    \hline
    \textbf{Dataset} & \textbf{Method} & \textbf{Architecture} & \textbf{Time-steps} & \textbf{Accuracy} & \textbf{Memory(GiB)} \\
    \hline
          & OTTT~\cite{xiao2022online}  & VGG(sWS) & 6     & 93.52\% & 4 \\
    \multirow{6}[3]{*}{CIFAR10} & S2A-STSU~\cite{tang2022snn2ann} & ResNet-17 & 5     & 92.75\% & 27.93 \\
          & IDE-LIF~\cite{xiao2021training} & CIFARNet-F & 30    & 91.74\% & 2.8 \\
          & Hybrid~\cite{rathi2020enabling} & VGG-16 & 100   & 91.13\% & 9.36 \\
          & Tandem~\cite{wu2021tandem} & CifarNet & 8     & 89.04\% & 4.2 \\
          & Skipper~\cite{singh2022skipper} & VGG-5 & 100   & 87.44\% & 4.6 \\
\cline{2-6}          & \textbf{RevSNN(Ours)} & ResNet-18 & 4     &   91.87\%    & \textbf{1.101} \\
    \hline
    \multirow{5}[4]{*}{CIFAR100} & IDE-LIF~\cite{xiao2021training} & CIFARNet-F & 30    & 71.56\% & 3.51{*} \\
          & OTTT~\cite{xiao2022online}  & VGG(sWS) & 6     & 71.05\% &  4.04{*} \\
          & S2A-STSU~\cite{tang2022snn2ann} & VGG-13 & 4     & 68.96\% & 31.05 \\
          & Skipper~\cite{singh2022skipper} & VGG-5  & 100   & 66.48\% & 4.6 \\
\cline{2-6}          & \textbf{RevSNN(Ours)} & ResNet-18 & 4     &   71.13\%    &\textbf{1.12}  \\
    \hline
    \multirow{8}[8]{*}{Tiny-ImageNet} & ND(Dense)~\cite{huang2023neurogenesis} & VGG-16 & 5     & 39.45\% & 3.99 \\
          & ND(90\% Sparsity)~\cite{huang2023neurogenesis} & VGG-16 & 5     & 39.12\% & 3.78 \\
          & ND(99\% sparsity)~\cite{huang2023neurogenesis} & VGG-16 & 5     & 33.84\% & 3.76 \\
\cline{2-6}           & \textbf{RevND(Ours)} & VGG-16 & 5     & 39.73\% & \textbf{2.01} \\
\cline{2-6}           & ND(Dense)~\cite{huang2023neurogenesis} & ResNet-19 & 5     & 50.32\% & 5.29 \\
          & ND(90\% Sparsity)~\cite{huang2023neurogenesis} & ResNet-19 & 5     & 49.25\% & 5.11 \\
          & ND(99\% sparsity)~\cite{huang2023neurogenesis} & ResNet-19 & 5     & 41.96\% & 5.09 \\
\cline{2-6}                    & \textbf{RevND(Ours)} & ResNet-19 & 5     & 50.63\% & \textbf{2.47} \\
    \hline
    \end{tabular}}%
  \label{tab:ComparisonOfSNNMethods}%
\end{table}%

Our approach (RevSNN) achieves a $\mathbf{2.54\times}$ memory reduction on the CIFAR10 dataset and a $\mathbf{3.13\times}$ memory reduction on the CIFAR100 dataset compared to the current SOTA SNN training memory-efficient method. At the same time, we also maintain a high level of accuracy. On the Tiny-ImageNet dataset, we only replaced the original SNN node with our designed reversible SNN node, keeping all other conditions consistent (RevND). As a result, the accuracy of our VGG-16 model structure is 0.28\% points higher than that of the original dense model and saves $\mathbf{1.87\times}$ more memory than the original work at 99\% sparsity. On the ResNet-19 model, our accuracy is 0.31\% points higher than the dense model, and saves $\mathbf{2.06\times}$ more memory than the original work at 99\% sparsity.

\subsection{Memory Consumption Evaluation} 
\label{Memory Consumption Evaluation}
To investigate whether our newly designed reversible SNN node achieves the expected memory savings compared to the original Spiking Neural Node, we incorporated our node into a range of architectures including VGG-11, 13, 16, 19, and ResNet-19, 34, 50, 101. For the VGG architectures, we examined the corresponding memory usage for timesteps ranging from 1 to 20, while for the ResNet architectures, we scrutinized the memory usage for timesteps from 1 to 10. These tests were conducted on the CIFAR-10 dataset. For all the experiments, we keep the batch size = 128.

\begin{figure}[htbp]
    \centering
    % \hspace*{-1.5cm}
    \includegraphics[width=0.99\textwidth]{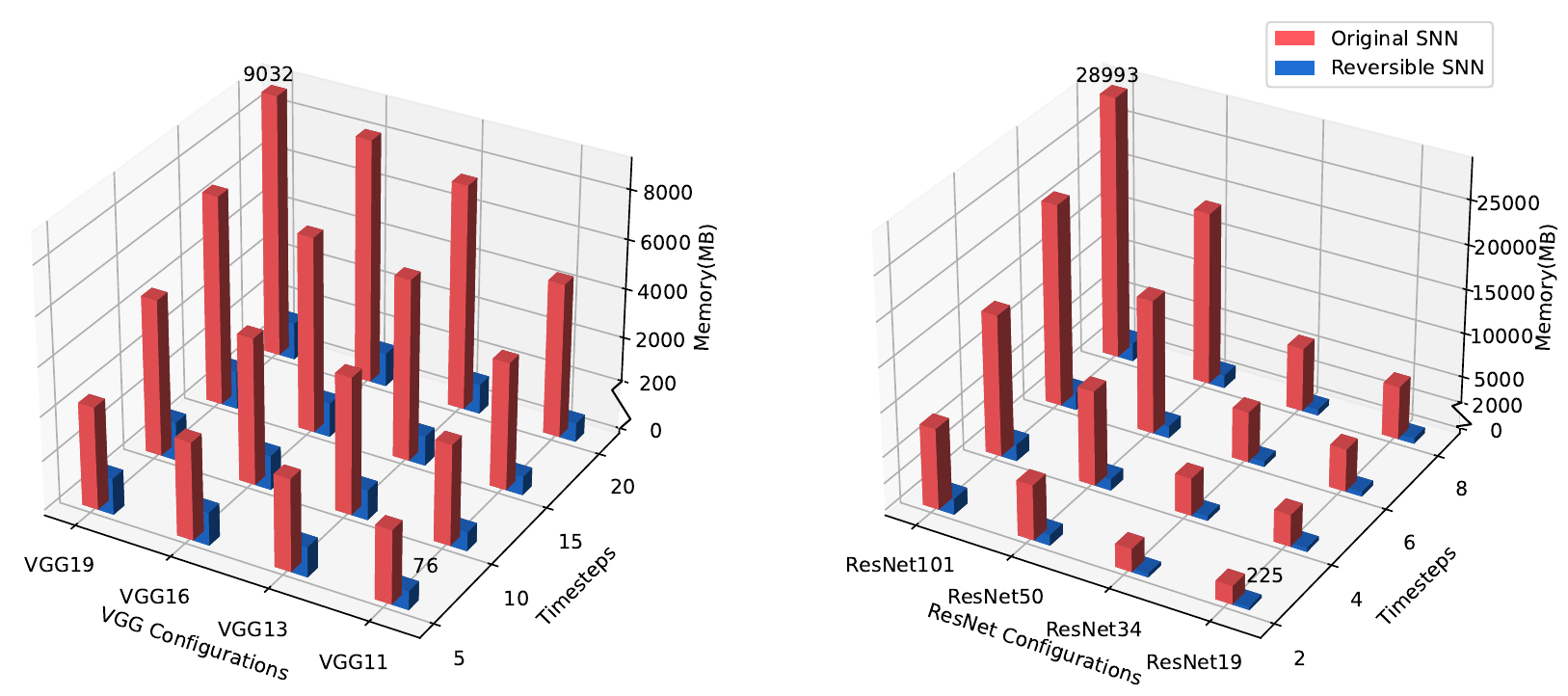} 
    \caption{Memory comparison between normal SNN node and our reversible SNN node.}
    \label{fig:experiment_memory}
\end{figure}

The SNN node memory comparison results is shown in Fig.\ref{fig:experiment_memory}. For VGG architecture, even when employing the most memory-intensive VGG-19 architecture with a timestep of 20, the cumulative memory usage for all the reversible SNN nodes within the entire network remains below 200MB. In contrast, using conventional SNN nodes demands a substantial amount of memory, up to 9032MB. For ResNet architectures, the ResNet-101 architecture with a timestep of 10 needs about 28993MB using conventional SNN node, but only 1382MB using our reversible SNN node. As the number of model layers and the timestep value increase, the memory savings achieved by our reversible SNN node become more pronounced. Specifically, when utilizing the VGG-19 architecture with a timestep of 20, our reversible SNN node enjoys a $\mathbf{58.65\times}$ memory reduction compared to the original SNN node. The specific data values are shown in the Appendix.

These experimental results align with our theoretical analysis in Section~\ref{memory analysis}, further validating that our design is able to significantly reduce memory usage.

\subsection{Training time Evaluation} 
\label{Training time Evaluation}
To investigate the time efficiency of our designed backpropagation architecture in comparison with the traditional reversible layer backpropagation method, we employ two sets of backpropagation architectures for our reversible SNN node. The first set utilizes the original reversible layer backpropagation method, while the second set incorporates our newly designed backpropagation architecture.

We employ VGG-11, VGG-13, VGG-16, and VGG-19 architectures with timesteps ranging from 1 to 10. We compare the time required for one iteration of training using the original SNN node, the reversible SNN node with the original reversible layer backpropagation method, and the reversible SNN node with our backpropagation architecture on the CIFAR-10 datasets. We perform all the experiments on an empty RTX6000 GPU and keep the batch size = 64. The reported times for each forward and backward pass are averages taken over all iterations within the first five epochs.

\begin{figure}[htbp]
    \centering
    % \hspace*{-1.5cm}
    \includegraphics[width=0.99\textwidth]{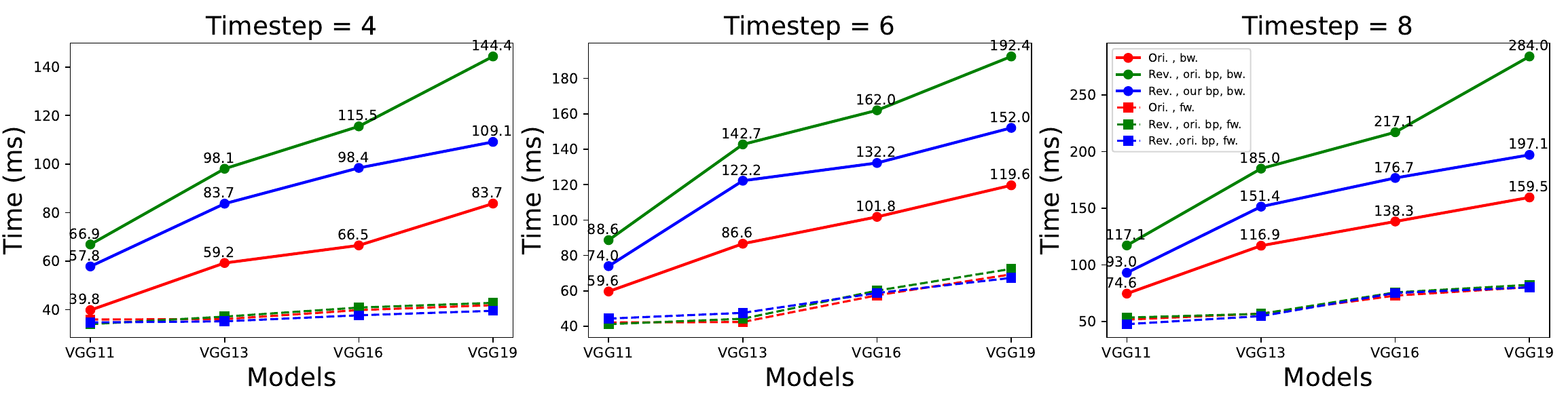} 
    \caption{Training time analysis. Solid lines: Backward process's duration; Dashed lines: Forward process's duration; \textcolor{red}{Red lines}: Training time for the original SNN; \textcolor{darkgreen}{Green lines}: Training time for the reversible SNN using the original reversible layer backpropagation method; \textcolor{blue}{Blue lines}: Training time for the reversible SNN employing our proposed backpropagation architecture.}
    \label{fig:time_analysis}
\end{figure}

Fig.\ref{fig:time_analysis} presents our measurement of the training time when the number of timesteps is set to 4, 6, and 8. 
% The solid lines represent the backward process's duration, while the dashed lines indicate the time consumed by the forward process. Each color corresponds to a specific training scenario: the red lines denote the time required for training the original SNN node, the green lines represent the time commitment for training the reversible SNN node using the original reversible layer backpropagation method, and the blue lines illustrate the training time for the reversible SNN node employing our proposed backpropagation architecture. 
The forward computation times for the three methods are virtually identical. The shortest backward processing time is exhibited by the original SNN node, primarily due to it records all intermediate values throughout the computation, thus eliminating the need for recalculations. Comparatively, among the two reversible SNN nodes, our backpropagation design achieves a $\mathbf{20\%}-\mathbf{30\%}$ increase in speed over previous reversible layer backpropagation method during the backward process. As the network expands, the superiority of our backpropagation design becomes increasingly evident. Specifically, under the VGG-19 architecture with a timestep of 8, our designed node is able to save $\mathbf{23.8\%}$ of the total training time compared to the reversible node using the original reversible layer backpropagation method. This aligns well with our previous theoretical predictions in Section~\ref{Inverse calculation of gradients}. Data for the other timesteps is shown in the Appendix.

\subsection{Ablation Study}
\noindent\textbf{Effects of parameters $\alpha$ and $\beta$ in our equations}

\begin{figure}[htbp]
    \centering
    % \hspace*{-1.5cm}
    \includegraphics[width=0.99\textwidth]{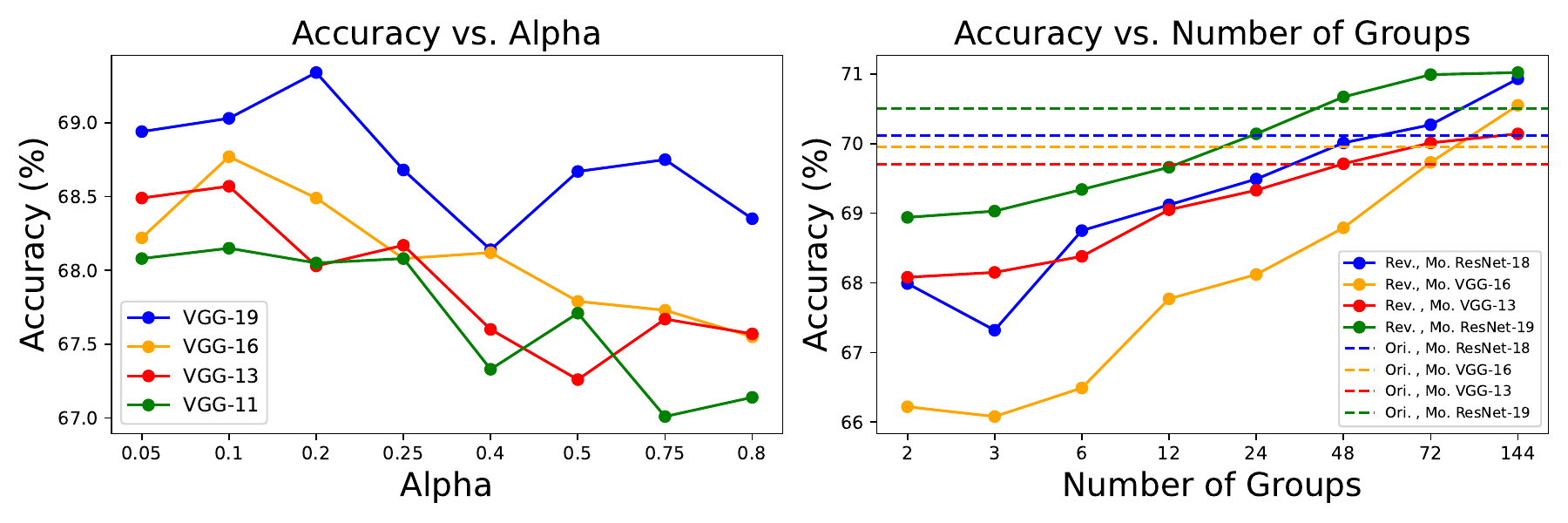} 
    \caption{\textbf{Left Figure}: Test VGG-19,VGG-16,VGG-13,VGG-11 models on CIFAR100 dataset by using different $\alpha$ settings. \textbf{Right Figure}: Change activations number from 128 to 144 for some fully connected layers inside ResNet-19, ResNet-18, VGG-16, VGG-13 and test model performance for different number of groups on CIFAR100. Rev.: Reversible SNN node. Ori.: Original SNN node. Mo.: Modified network (Change some fully connected layers).}
    \label{fig:ablation study}
\end{figure}

In Eq.\ref{eq:forward Y0} and Eq.\ref{eq:forward V0}, we have two parameters: $\alpha$ and $\beta$. The optimal setting for the parameter $\beta$ is 1, as this maximizes the preservation of the original features of the data. We conduct experiments to assess the impact of the $\alpha$ parameter on the model's performance. We vary the $\alpha$ parameter from 0.05 to 0.8, and then employ architectures VGG-19, VGG-16, VGG-13, and VGG-11 to evaluate the accuracy on the CIFAR100 dataset. The results are shown on the left of Fig.\ref{fig:ablation study}. We observe that varying $\alpha$ within the range of 0.05 to 0.8 impacts the final accuracy by approximately 1\%. Generally, the model exhibits optimal performance when $\alpha$ is set between 0.1 to 0.2.

\noindent\textbf{Effects of number of groups for the various states}

In Section \ref{sec:Reversible SNN forward calculation}, 
We introduce a method of splitting various input states into two groups along the last dimension. Nonetheless, this method might encounter issues under specific circumstances. For instance, if the last dimension of a tensor is an odd number, it cannot be evenly divided into two groups. To address this, we enhance the original algorithm: we divide the various input states into $n$ groups according to the number of elements $n$ in the last dimension. Eq.\ref{eq:forward M0}-\ref{eq:forward V0} is then executed sequentially for each group. This enhancement further improves the universality of our algorithm. 

To evaluate the impact of the number of groups on the model, we modified part of the fully connected layers in the original ResNet-19, ResNet-18, VGG-16, VGG-13  network from 128 activations to 144 activations. This is to allow it to have a wider variety of factors. We then evaluate the model's performance with the number of groups set to 2, 3, 6, 12, 24, 48, 72, and 144 respectively on CIFAR100 dataset. The results are shown on the right of Fig.\ref{fig:ablation study}. 
We observe that the training accuracy improves as the number of groups increases. When the number of groups approaches the number of elements $n$ in the last dimension, the accuracy typically surpasses that of the original SNN node. This is attributed to a larger number of groups yielding a higher fidelity representation of the original data.

\section{Conclusion and Discussion}

This work addresses a fundamental bottleneck of current deep SNNs: their high GPU memory consumption. We have designed a novel reversible SNN node that is able to reduce memory complexity from $\mathcal{O}(n^2)$ to $\mathcal{O}(1)$. Specifically, our reversible SNN node allows our SNN network to achieve $\mathbf{2.54}$ times greater memory efficiency than the current SOTA SNN memory-efficient work on the CIFAR10 dataset, and $\mathbf{3.13}$ times greater on the CIFAR100 dataset. Furthermore, in order to tackle the prolonged training time issue caused by the need for recalculating intermediate values during backpropagation within our designed reversible SNN node, we've innovated a new backpropagation approach specifically suited for reversible architectures. This innovative method, when compared to the original reversible layer architecture, achieves a substantial reduction in overall training time by $\mathbf{23.7\%}$.
As a result, we are able to train over-parameterized networks that significantly outperform current models on standard benchmarks while consuming less memory.
% \section*{Acknowledgments}

% We would like to express our gratitude to all those who have contributed to this research. First and foremost, we thank our colleagues and mentors for their valuable insights and constructive feedback throughout the project. We also appreciate the support from the funding agencies, which have enabled us to carry out this work.

% Additionally, we acknowledge the authors of the previous studies and tools that have been instrumental in shaping our research direction. Lastly, we extend our thanks to the reviewers for their time and valuable suggestions, which have helped improve the quality of our work.

\bibliography{bio}
\bibliographystyle{plain}
% \bibliographystyle{apalike} 

% \appendix
% \input{Sections/Appendix}

\end{document}